\documentclass[sigconf]{acmart}
\AtBeginDocument{%
  }

\setcopyright{acmcopyright}
\copyrightyear{2023}
\acmYear{2023}
\acmDOI{XXXXXXX.XXXXXXX}

\acmPrice{15.00}
\acmISBN{978-1-4503-XXXX-X/18/06}

\usepackage{algorithm}
\usepackage{algorithmic}

\usepackage{multirow}

\usepackage{pifont}

\acmSubmissionID{553}
\renewcommand\footnotetextcopyrightpermission[1]{}
\begin{document}

\title{Adversarial Infrared Blocks: A Black-box Attack to Thermal Infrared Detectors at Multiple Angles in Physical World}


\author{Chengyin Hu}
\affiliation{%
  \institution{University of Electronic Science and Technology of China}
  \city{Chengdu}
  \country{China}}
\email{cyhuuesct@gmail.com}

\author{Weiwen Shi}
\affiliation{%
  \institution{University of Electronic Science and Technology of China}
  \city{Chengdu}
  \country{China}}
\email{weiwen_shi@foxmail.com}

\author{Tingsong Jiang $^{\ast}$}
\affiliation{%
  \institution{Chinese Academy of Military Science}
  \city{Beijing}
  \country{China}}
\email{tingsong@pku.edu.cn}

\author{Wen Yao $^{\ast}$}
\thanks{$\ast$ Corresponding author}
\affiliation{%
  \institution{Chinese Academy of Military Science}
  \city{Beijing}
  \country{China}}
\email{wendy0782@126.com}

\author{Ling Tian}
\affiliation{%
  \institution{University of Electronic Science and Technology of China}
  \city{Chengdu}
  \country{China}}
\email{lingtian@uestc.edu.cn}

\author{Xiaoqian Chen}
\affiliation{%
  \institution{Chinese Academy of Military Science}
  \city{Beijing}
  \country{China}}
\email{chenxiaoqian@nudt.edu.cn}

\begin{teaserfigure}
  \includegraphics[width=\textwidth]{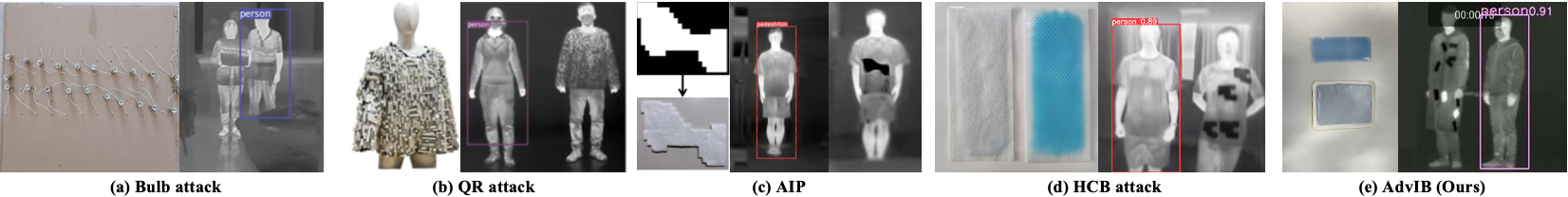}
  \caption{Demonstration of our proposed AdvIB and other physical infrared attacks.}.
  \label{figure1}
\end{teaserfigure}

\begin{abstract}
 Thermal infrared detectors have a vast array of potential applications in pedestrian detection and autonomous driving, and their safety performance is of great concern. Recent works use bulb plate, "QR" suit, and infrared patches as physical perturbations to perform white-box attacks on thermal infrared detectors, which are effective but not practical for real-world scenarios. Some researchers have tried to utilize hot and cold blocks as physical perturbations for black-box attacks on thermal infrared detectors. However, these attempts have not yielded robust and multi-angle physical attacks, indicating limitations in the approach. To overcome the limitations of existing approaches, we introduce a novel black-box physical attack method, called adversarial infrared blocks (\textbf{AdvIB}). By optimizing the physical parameters of the infrared blocks and deploying them to pedestrians from multiple angles, including the front, side, and back, AdvIB can execute robust and multi-angle attacks on thermal infrared detectors. Our physical tests show that the proposed method achieves a success rate of over 80\% under most distance and angle conditions, validating its effectiveness. For stealthiness, our method involves attaching the adversarial infrared block to the inside of clothing, enhancing its stealthiness. Additionally, we perform comprehensive experiments and compare the experimental results with baseline to verify the robustness of our method. Overall, our proposed AdvIB method offers a promising avenue for conducting robust and multi-angle black-box attacks on thermal infrared detectors, with potential implications for real-world security applications.
\end{abstract}



\keywords{Thermal infrared detectors; Adversarial infrared blocks; Effectiveness; Stealthiness; Robustness; Multi-angle black-box attacks}

\maketitle

\section{Introduction}
The utilization of thermal infrared detectors in pedestrian detection and automatic driving displays considerable potential.
Nonetheless, it should be noted that the implementation of such detectors carries certain security risks. In-depth investigations conducted in this field \cite{ref1,ref2,ref3,CVPR2023} have revealed that these systems are susceptible to physical attacks, which can potentially jeopardize the safety of both pedestrian detection and autonomous driving.


\begin{figure*}
\centering
\setlength{\abovecaptionskip}{-1pt}
\includegraphics[width=0.8\linewidth]{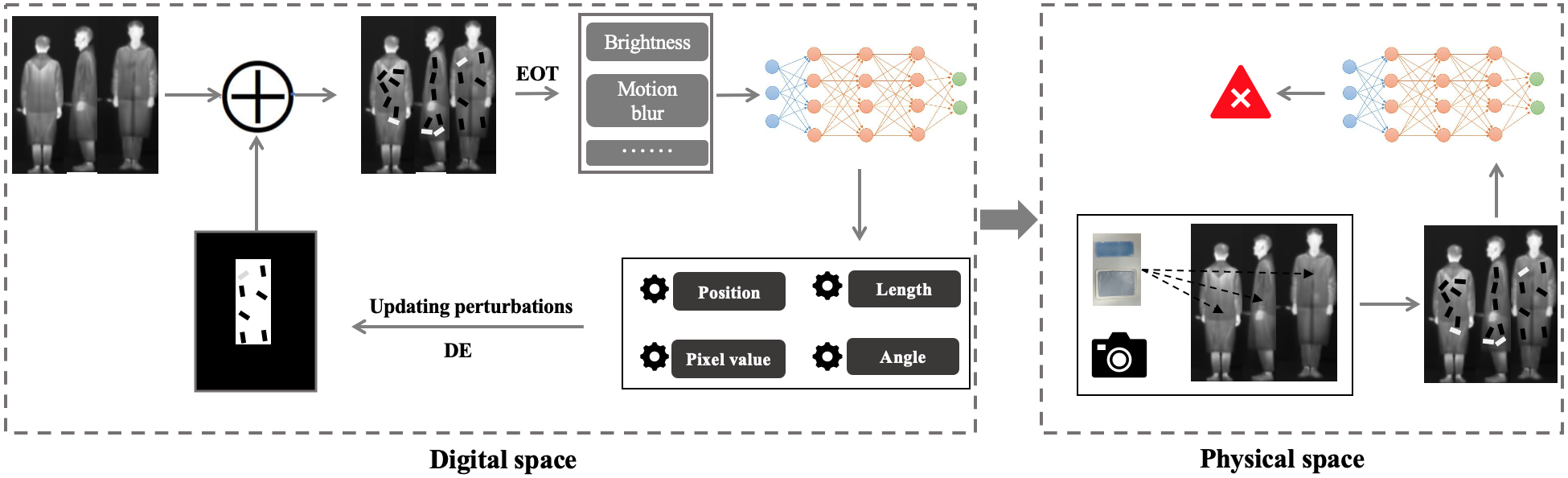} 
\caption{Overview of our work. The left depicts the process of optimizing the adversarial infrared blocks in a digital environment, while the right demonstrates the process of executing the adversarial attack in a physical environment.}.
\label{figure2}
\end{figure*}

The thermal infrared detector is a novel detector that leverages deep neural networks (DNNs) in conjunction with infrared imaging technology. However, the security of DNNs has become a focal point of interest for many researchers. Szegedy et al. \cite{ref4} were among the first to identify that fine-tuned DNNs are susceptible to minor perturbations, known as adversarial perturbations. Presently, most adversarial attacks are focused on visible images \cite{ref5,ref6,ref7,ref8, ref52}. \textbf{1)} Digital attacks, in particular, add subtle perturbations to the input image, which are imperceptible to the human eye. \textbf{2)} Physical attacks \cite{ref51,ref9,ref10,ref53}, on the other hand, necessitate the capturing of adversarial samples through a camera, followed by feeding them into deep neural networks for attack. Consequently, physical perturbations need to be designed with relatively larger magnitudes to ensure their detectability by the camera.

Some researchers have utilized small bulbs \cite{ref1} to launch physical attacks against infrared pedestrian detectors, but such bulbs are highly conspicuous in reality when held by pedestrians. Some research studies have utilized thermal insulation materials to create "QR codes" \cite{ref2} and irregular patches \cite{CVPR2023} as a means of attacking infrared pedestrian detectors. However, these materials are often conspicuous, and covering them on pedestrians can draw unwanted attention. In another study \cite{ref3}, hot and cold blocks were used to create nine grids that were affixed to the inside of clothing to execute black-box attacks. However, this method cannot achieve robust and multi-angle adversarial effects.

Based on the aforementioned discussion, we introduce a black-box physical attack on infrared pedestrian detectors, which called adversarial infrared blocks. This method optimizes the physical parameters of infrared block (such as position, angle, length, and color) to execute the attack. In comparison to the Bulb attack \cite{ref1}, QR attack \cite{ref2}, and Adversarial Infrared Patches (AIP for short) \cite{CVPR2023}, our approach is more suitable for real-world scenarios. Furthermore, our approach attaches infrared blocks to the inside of clothing, making it stealthy. Compared with HCB \cite{ref3}, our method is more robust and applicable due to the flexible design and global optimization adopted in our method. Figure \ref{figure1} and Table \ref{T} respectively provide a visual comparison and the comparison of similarities and differences between our approach and existing methodologies. 

\begin{table} 
	\centering
    \caption{The comparison between existing methods and our method.}.
    \label{T}
	\begin{tabular}{cccc}

    \hline
    Method & Multi-angle & Scenario & Stealthiness\\
    \hline
    Bulb attack \cite{ref1} & \ding{55} & White-box & \ding{55}\\
    \hline
    QR attack \cite{ref2} & \ding{51} & White-box & \ding{55}\\
    \hline
    AIP \cite{CVPR2023} & \ding{55} & White-box & \ding{55}\\
    \hline
    HCB \cite{ref3} & \ding{55} & Black-box & \ding{51}\\
    \hline
    AdvIB(Ours) & \ding{51} & Black-box & \ding{51}\\
    \hline

\end{tabular}
\vspace{-0.2cm}
\end{table}

In order to generate perturbation patterns on an infrared camera, we utilized physical perturbations in the form of hot blocks (Warming pastes) and cold blocks (Cooling pastes), where the hot block and cold block correspond to white and black patterns, respectively, in the resulting infrared image. These blocks are attached to the inside of a pedestrian's clothing, making the perturbation difficult for the human eye to detect, but detectable by infrared cameras. Our design is thus wearable and stealthy, easily deployable in the physical world. Firstly, we use the differential evolution algorithm [15] for optimization to determine the physical parameters of infrared blocks with the most adversarial effect. Secondly, based on these parameters, we paste hot and cold blocks on the cloth of pedestrians to carry out physical attacks against the infrared pedestrian detector. It is worth noting that our physical attack costs less than 5\$, making our approach much more accessible. Figure \ref{figure2} illustrates our approach. Our major contributions are summarized below:

\begin{itemize}
\item We propose a black-box physical attack against infrared pedestrian detector, AdvIB, which has the advantages of remarkable adversarial effect, multi-angle attack, better stealthiness, wearability and low cost, making it a security hazard that cannot be ignored in the physical world.

\item We model the proposed AdvIB and evaluate it on current advanced target detectors. Comprehensive experimental results demonstrate the effectiveness, stealthiness, and robustness of AdvIB in both digital and physical environments.

\item We summarize the experimental results and compare them with baseline to verify the superiority of our approach. we also look forward to the idea of physical attacks against infrared detection systems.
\end{itemize}

\section{Related works}

\subsection{Adversarial attacks in the visible light field}

Currently, most adversarial attacks are predominantly deployed in the visible light field \cite{ref16,ref17,ref18,ref19}. Szegedy et al. \cite{ref4} first discovered that well-trained DNNs are susceptible to slight perturbations, leading to a multitude of related studies \cite{ref20,ref21,ref22}. Most researchers \cite{ref23,ref24,ref25} have utilized ${L}_{2}$ and ${L}_{\infty}$ norms to restrict the perturbations to imperceptible levels for human observers. Some recent works have also focused on other image attributes, such as color, texture, and camouflage, to generate attacks \cite{ref26,ref27,ref28,ref29,ref30}. Additionally, some scholars \cite{ref31,ref32} have utilized digital synthesis methods to simulate raindrops to deploy adversarial attacks, subsequently using these generated adversarial samples to improve the model's robustness. These works have been primarily conducted in digital environments, where attackers can directly modify input images to generate adversarial samples. In contrast, physical attacks require the adversarial sample to be captured by a camera and then fed into the target model to perform an attack. Traditional physical attacks use optimized stickers, posters, etc., as physical perturbations \cite{ref33,ref34,ref35,ref36}, which are then pasted on the target objects to achieve an effective physical attack.
Recently, many researchers have shifted their focus towards light-based \cite{ref37,ref38} and camera-based \cite{ref39,ref40} attacks. These attacks avoid directly modifying the target objects and are relatively stealthier compared to traditional physical attacks.

\subsection{Adversarial attacks in the infrared field}

Zhu et al. \cite{ref1} conducted the first study on physical attacks in the infrared domain and proposed the bulb attack by using a set of bulb panels to deceive well-trained infrared pedestrian detectors. The mechanism involves utilizing the heat source from the bulbs, which generates a white perturbation pattern under the infrared camera, to deceive the infrared pedestrian detector. However, the need for the pedestrian to hold the bulb panels is not stealthy, and deploying physical attacks at various angles is impractical. In a subsequent study, Zhu et al. \cite{ref2} investigated using thermal insulation materials to obstruct the body heat source for physical attacks against the infrared pedestrian detector and proposed a QR attack. They employed aerogel as insulation to optimize adversarial QR code patterns in a digital environment and then implemented the infrared adversarial suit through manual simulation. Their experiments demonstrated the effectiveness of attacking the infrared pedestrian detector from different angles. However, the manual simulation of "QR" patterns is costly, and the generated infrared suit is conspicuous and suspicious to human observers. Wei et al. \cite{CVPR2023} also used thermal insulation to generate irregular patches to perform physical infrared attacks, however this method is not capable of performing physical attacks from various angles. To achieve greater stealth, Wei et al. \cite{ref3} introduced HCB attack. They used hot and cold blocks as physical perturbations, by employing the Particle Swarm Optimization (PSO) \cite{PSO} algorithm to optimize parameters and perform black-box attacks against the infrared pedestrian detector. As the hot and cold blocks are concealed inside clothing, it is difficult for human eyes to detect their presence. Nevertheless, the nine-grid design of the perturbation pattern used in HCB only allows for frontal physical attacks, limiting its practical use against attacks from different angles. Simultaneously, the nine-grid design presents difficulties in achieving global optimization, limiting their ability to produce a stronger adversarial effect.

Our AdvIB method presents a novel approach for performing black-box physical attacks against fine-turned infrared pedestrian detectors. It utilizes hot and cold blocks as perturbations, and by optimizing their physical parameters and deploying them from various angles, it is capable of executing multi-angle attacks, rendering it more versatile than the Bulb attack \cite{ref1} and AIP \cite{CVPR2023}. Moreover, our method is more covert in deploying perturbations than the QR attack \cite{ref2} and is simpler to operate. In comparison to the HCB \cite{ref3}, our approach uses fewer perturbations and is more effective and robust. The HCB technique is incapable of executing attacks from various angles, whereas our AdvIB method can, thus making it more practical in physical attacks.

\begin{figure}
\centering
\setlength{\abovecaptionskip}{-1pt}
\includegraphics[width=1\columnwidth]{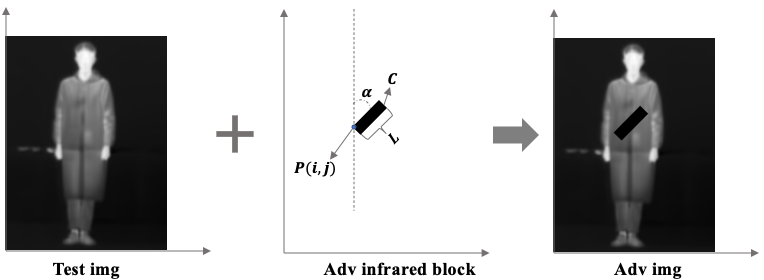} 
\caption{Infrared block modeling. AdvIB utilizes physical parameters of the most adversarial infrared blocks to perform attacks. It can be seen that our modeling methodology enables flexible adjustment to infrared blocks.}.
\vspace{-0.5cm}
\label{figure3}
\end{figure}

\section{Method}

\subsection{Problem definition}
Given an infrared pedestrian dataset $DS$, let $X$ and $Y$ denote the sets of clean images and their corresponding ground truth labels, respectively, $f$ represents a target detector. Therefore, for $X \in DS$, the target detector's pre-training model $f: X \rightarrow Y$ can predict the labels $y$ that matches the ground truth label $Y$, where $y$ contains the boundary box position information ${V}_{pos}$, target probability ${V}_{obj}$, and target category ${V}_{cls}$:

\begin{equation}
    \label{Formula 1}
    y:=[{V}_{pos},{V}_{obj},{V}_{cls}]=f(X)
\end{equation}

\subsection{Infrared block modeling}

The proposed AdvIB is designed to employ infrared blocks, such as hot and cold blocks, to deceive the infrared pedestrian detector, which is a patch-based attack. Our modeling of the infrared block is demonstrated in Figure \ref{figure3}, where four physical parameters are utilized to represent an infrared block. These parameters consist of position, pixel value, length, angle, which are defined as follows: 

\textbf{Position $P$:} To define the position of each infrared block, we utilize $P(i,j)$ to represent the top-left point coordinates of the block.  In order to achieve a more potent attack, we use multiple blocks to launch the attack, denoted as $P=\{{P}_{1}({i}_{1},{j}_{1}),…,{P}_{k}({i}_{k},{j}_{k})\}$, where $k$ is the number of infrared blocks. The number of infrared blocks utilized directly impacts the strength of the adversarial effect and the stealthiness of the attack. Thus, we conduct a series of ablation experiments to tradeoff the effectiveness and stealthiness. Given $\mathcal{M}$ as a mask used to locate pedestrians, the location information of infrared blocks is represented by $P \cap \mathcal{M}$.

\textbf{Pixel value $C$:} The pixel value of an infrared block is represented by $C$. In grayscale, there is only a single-color channel with a range of 0 to 255, where 0 and 255 correspond to black and white, respectively. In order to represent the grayscale imaging effect of the infrared block, we map the value of $C$ to a number between 0 and 1 with a spacing of 0.1. It should be noted that the hot and cold blocks captured by the infrared camera only exhibit grayish white and grayish black imaging effects, respectively. Since the physical perturbation used in our experiments cannot adjust the temperature, it is not possible to achieve different grayscale perturbation effects under the infrared camera.  Therefore, we simulate the infrared blocks with $C=0.9$ (hot block) and $C=0.1$ (cold block).

\textbf{Length $L$:} The parameter $L$ denotes the ratio of the length of the infrared block to the width of the bounding box. To simulate the actual physical perturbation pattern more accurately, we set the width $W$ of the infrared block to $0.74L$ when $C=0.9$ and $0.45L$ when $C=0.1$, respectively.

\textbf{Angle $\alpha$:} $\alpha$ represents the angle between the long side of the infrared block and the vertical axis of the image. Once the position $P$, length $L$, and angle $\alpha$ of the infrared block are determined, the position coordinates of the other three points can be calculated. This enables us to explore various attack scenarios by combining different positions and angles. Here, $\alpha \in [{0}^{\circ}, {180}^{\circ}]$.

\subsection{Infrared blocks attack}
We use $\theta=\{{B}_{1},{B}_{2},...,{B}_{k}\}$ to represent the generated infrared blocks, where ${B}_{1}=({P}_{1},{C}_{1},{L}_{1},{\alpha}_{1}),...,{B}_{k}=({P}_{k},{C}_{k},{L}_{k},{\alpha}_{k})$, ${\theta}$ is subject to predefined constraints, ${\theta}_{min}$ and ${\theta}_{max}$, which are adjustable. The process for generating adversarial samples using the above-defined parameters can be expressed as follows:

\begin{equation}
    \label{Formula 2}
    {X}_{adv} = S(X, \theta, \mathcal{M}) \quad \theta \in ({\vartheta}_{min},{\vartheta}_{max})
\end{equation}
where $S$ represents a simple linear fusion approach to synthesize the generated simulation infrared blocks and clean sample $X$, so as to obtain the adversarial sample ${X}_{adv}$ in the digital environment.

In our physical experiment, we affix hot and cold blocks to the front, back, and side of the pedestrian to achieve the desired adversarial effect against the infrared pedestrian detector from various angles. To enhance the robustness of our proposed method against different changes, we employ Expectation over Transformation (EOT) \cite{ref50}, an effective tool for dealing with domain shifts between digital and physical domains. EOT involves applying a distribution of transforms, $\mathcal{T}$, to simulate domain shifts. Each instance of $\mathcal{T}$ comprises a set of random image transformations, including view conversion, brightness adjustment, and downsampling, etc. Importantly, $\mathcal{T}$ can solve the problem of slight errors in the pixel value and position of the simulated infrared block. Thus, the adversarial sample in the physical domain can be expressed as:

\begin{equation}
    \label{Formula 4}
    {X}_{adv} = {\mathbb{E}}_{t \sim \mathcal{T}}(t(S(X, \theta, \mathcal{M}),\theta)) \quad \theta \in ({\vartheta}_{min},{\vartheta}_{max})
\end{equation}


The objective of our research is to determine the optimal physical parameters $\theta$ of the infrared blocks, which is used to generate adversarial sample ${X}_{adv}$, so as to render the pedestrian undetectable. To increase the realism of our approach, we consider a black-box attack scenario, wherein we lack detailed information regarding the model's architecture and parameters. Our access is limited to only the input image and the detection information of the model output: ${V}_{pos}$, ${V}_{obj}$, ${V}_{cls}$. Consequently, we use the probability ${V}_{obj}$ as the adversarial loss and formulate the optimization objective as minimizating ${V}_{obj}$:

\begin{equation}
    \label{Formula 5}
    \mathop{\arg\min}_{\theta}{\mathbb{E}}_{t \sim \mathcal{T}}({V}_{obj} \leftarrow t(f({X}_{adv}),\theta))
\end{equation}

To achieve faster global optimization, we employ the differential evolution (DE) algorithm \cite{ref15} to optimize the proposed objective function. DE is a population-based heuristic search algorithm that is known for its efficiency in global optimization. The algorithm utilizes swarm intelligence generated by mutual cooperation and competition among individuals within a group to guide the direction of optimization. The following will describe the process and algorithm of using DE to optimize our method:

\textbf{Initiation.} We initiate the optimization process with a randomly generated initial population:

\begin{equation}
    \label{Formula 7}
    POP=[{\theta}_{1},{\theta}_{2},...,{\theta}_{G}]
\end{equation}
where $G$ denotes the population size, ${\theta}_{g}$ ($g=1,2,...,G$) denotes a candidate solution in $POP$.

\textbf{Mutation.} For each individual $\theta$ in the population, we obtain the mutated individual by adding it to the vector difference of two randomly selected individuals in the population:

\begin{equation}
\centering
\begin{split}
    \label{Formula 8}
    &\quad {\theta}_{g1} =  Clipping\{{\theta}_{g1} + {R}_{m}({\theta}_{g2} - {\theta}_{g3})\} \\
    &s.t. \quad g1, g2, g3 \in [1, 2,..., G] \quad g1\neq g2\neq g3
\end{split}
\end{equation}
$Clipping{}$ indicates that parameters outside the constraint range are randomly clipped to the constraint range. ${R}_{m}$ represents the mutation rate of DE.

\begin{algorithm}[t]
	\renewcommand{\algorithmicrequire}{\textbf{Input:}}
	\renewcommand{\algorithmicensure}{\textbf{Output:}}
	\caption{Pseudocode of AdvIB}
	\label{algorithm1}
	\begin{algorithmic}[1]
	
		\REQUIRE Clean image $X$, Detector $f$, Population size $G$, Iterations $Step$, Mutation rate ${R}_{m}$, Crossover rate ${R}_{c}$;

        \ENSURE Physical parameters ${\theta}^{\star}$;

        \STATE \textbf{Initialize} $POP=[{\theta}_{1},{\theta}_{2},...,{\theta}_{G}]$, ${V}^{\star}$, ${\theta}^{\star}$;
        
        \FOR{$steps$ $\leftarrow$ 0 to $Step$}
            \STATE ${POP}_{\mathcal{P}}=POP$;
        \FOR{each $\theta$ in $POP$}
            \STATE $\theta \leftarrow Mutation(POP,{R}_{m})$;
            \STATE $\theta \leftarrow Crossover(POP,{R}_{c})$;
            \STATE ${X}_{adv} = {\mathbb{E}}_{t \sim \mathcal{T}}(t(S(X, \theta, \mathcal{M}),\theta))$;
            \STATE $[{V}_{pos},{V}_{obj},{V}_{cls}] \leftarrow f({X}_{adv})$;
        \IF{${V}_{obj}<{V}^{\star}$}
            \STATE ${V}^{\star}={V}_{obj}$;
            \STATE ${\theta}^{\star}=\theta$;
        \ENDIF
        \ENDFOR
            \STATE $POP=Selection({POP}_{\mathcal{P}},POP)$;
        \ENDFOR
        \STATE Output ${\theta}^{\star}$;
	
	\end{algorithmic}  
\end{algorithm}

\textbf{Crossover.} For each component ${\theta}_{g}^{d}$ in the individual ${\theta}_{g}$, we generate a random number between 0 and 1.  When the number is less than or equal to ${R}_{c}$, we perform binomial crossing, which can be represented as follows:

\begin{equation}
\centering
\begin{split}
\centering
    \label{Formula 9}
    &{\theta}_{g}^{d}=
        \begin{cases}
        {\theta}_{g}^{\mathcal{P},d} & rand[0,1] \leqslant {R}_{c} \\
        {\theta}_{g}^{d} & otherwise
        \end{cases}\\
    &s.t. \quad g \in [1, 2,..., G] \quad d \in [1,2,...,\mathcal{D}]
\end{split}
\end{equation}
where ${\theta}_{g}^{d}$ represents the $d$-th component of ${\theta}_{g}$, ${\theta}_{g}^{\mathcal{P},d}$ represents the $d$-th component in the parent of ${\theta}_{g}$, $\mathcal{D}$ denotes the dimension of ${\theta}_{g}$, $rand[0,1]$ represents the random number between 0 and 1, and ${R}_{c}$ represents the crossover rate of DE.

\textbf{Selection.} The new generation population is selected by comparing the fitness of each individual in the current population with that of the parent population. In this work, we use the target probability value detected by the detector as the fitness value. Therefore, the lower the target probability is, the more adversarial the individual is:

\begin{equation}
\centering
\begin{split}
    \label{Formula 10}
    &{\theta}_{g}=
        \begin{cases}
        {\theta}_{g}^{\mathcal{P}} & {V}_{obj} > {V}_{obj}^{\mathcal{P}} \\
        {\theta}_{g} & otherwise
        \end{cases}\\
    &s.t. \quad g \in [1, 2,..., G]
\end{split}
\end{equation}
${V}_{obj}$ and ${V}_{obj}^{\mathcal{P}}$ are the fitness values of ${\theta}_{g}$ and ${\theta}^{\mathcal{P}}$, respectively (where ${\theta}_{g}^{\mathcal{P}}$ is the individual corresponding to ${\theta}_{g}$ in the parent population).



The proposed AdvIB employs DE optimization to obtain the most adversarial physical parameters of the infrared blocks and generate adversarial samples capable of fooling the detector, then uses EOT to further improve the robustness of the adversarial samples. Algorithm \ref{algorithm1} shows the pseudocode of AdvIB, which takes clean sample $X$, detector $f$, population size $G$, iteration number $Step$, mutation rate ${R}_{m}$ and crossover rate ${R}_{c}$ as inputs. The algorithm initializes the population randomly and conducts mutation, clipping, and crossover operations for each individual. The algorithm outputs the physical parameter ${\theta}^{\star}$, which is used to perform subsequent physical attack experiments. 

\begin{figure}
\centering
\setlength{\abovecaptionskip}{-1pt}
\includegraphics[width=0.8\columnwidth]{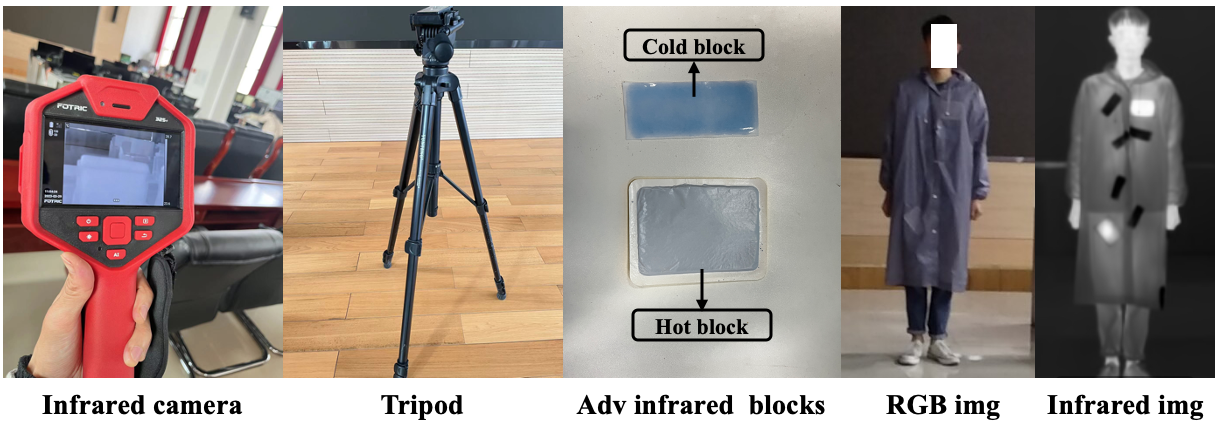} 
\caption{Experimental devices. In this experiment, we use a FOTRIC 325+ infrared camera, a tripod, and some hot and cold blocks as experimental devices.}.
\vspace{-0.6cm}
\label{figure4}
\end{figure}

\section{Experiments}

\subsection{Experimental settings}
\textbf{Dataset.}The FLIR\_ADAS v1\_3 dataset provided by FLIR is utilized in this study. The dataset comprises of 10,228 infrared images, which were extracted from short videos and a continuous video, the FLIR Tau2 infrared camera (13 mm f/1.0, 45-degree HFOV and 37-degree VFOV, FPA640 $\times$ 512, NETD < 60mK) was used to capture these images. The annotations for the dataset were manually created and contain four object classes: people, bicycle, car, and dog. Our proposed AdvIB method is designed specifically for infrared pedestrian detectors. To this end, we filter the original dataset and only retain images containing pedestrians that are taller than 120 pixels, resulting in a subset of 1011 images. This filtered dataset is named FLIR-InfraredBlocks (FLIR-IB). For our experiments, 710 images from FLIR-IB are utilized for training, while the remaining 301 images are set aside for testing.

\textbf{Target detector.} We employ Yolo v3 \cite{ref48} as the target detector to train the infrared pedestrian detector. To conform to the model specifications, we resize the image to 416 $\times$ 416, we use the pre-training weights of Yolo v3 and then fine-turn on FLIR-IB. The resulting infrared pedestrian detector achieves AP scores of 90.1\% and 90.7\% on the training and test sets of FLIR-IB, respectively.

\textbf{Experimental devices.} For the physical experiments, the experimental devices utilized in this study is depicted in Figure \ref{figure4}, including a FOTRIC 325+ infrared camera, a tripod, hot and cold blocks. The FOTRIC 325+ camera is equipped with 352 $\times$ 264 FPA, NETD < 40mk, and 704 $\times$ 528 SR. The infrared camera supports 16 color palettes, including Grey, Iron10, Iron, etc. The hot blocks are capable of maintaining an average temperature of ${53}^{\circ}C$ for 6 hours, while the cold blocks could be cooled to ${24}^{\circ}C$ for 4 hours.

\textbf{Evaluation metrics.} Our proposed AdvIB aims to conceal pedestrians from infrared pedestrian detectors. As such, the average precision (AP) metric is used to evaluate the performance of the infrared pedestrian detector under the adversarial dataset. It is worth noting that a lower AP score indicates better adversarial efficacy. Moreover, the attack success rate (ASR) is employed to evaluate the effectiveness of AdvIB, which is defined as follows:

\begin{equation}
\label{eq:Positional Encoding}
\begin{split}
    &{\rm ASR}(X) = 1-\frac{1}{N}\sum_{i=1}^{N}F({label}_{i})\\
    &F({label}_{i})=
        \begin{cases}
        1 & {label}_{i} \in {L}_{pre} \\
        0 & otherwise
        \end{cases}
\end{split}
\end{equation}
where $N$ represents the number of true positive labels detected by the detector in the dataset $DS$ in the absence of attack, and ${L}_{pre}$ is the set of all labels detected under attack. A higher ASR indicates a more effective attack.

\textbf{Competing methods.} We conduct a comparative study to assess the efficacy of AdvIB in the realm of infrared attacks relative to baseline methods. Specifically, we evaluate the stealthiness of AdvIB in comparison to the Bulb attack \cite{ref1}, QR attack \cite{ref2}, AIP \cite{CVPR2023} and HCB \cite{ref3}. We only compare the adversarial effect of the proposed method with HCB, given that HCB and AdvIB performs black-box attacks while the other methods are white-box attacks.

\textbf{Other details.} To approximate the physical perturbation pattern, the pixel values for the hot and cold blocks were set to $C=0.9$ and $C=0.1$, respectively. The DE algorithm parameters are set as follows: $G=100$, $Step=10$, ${R}_{m}=0.5$, ${R}_{c}=0.6$. All attack experiments are carried out on a NVIDIA GeForce RTX 4090 GPU device.

\subsection{Evaluation of effectiveness}

\textbf{Digital test.} We conduct a series of ablation experiments to assess the effectiveness of AdvIB in a digital environment. Our aim is to identify a reasonable attack configuration based on the experimental results. Our initial focus is on the number and length of infrared blocks, as these are key factors that influence the effectiveness of the attack. Insufficient infrared blocks resulted in weak adversarial effects, while an excessive number of blocks reduced the stealthiness of the attack. shorter side lengths of infrared blocks resulted in weaker adversarial effects, while longer side lengths reduced the stealthiness of the attack. To strike a balance, we vary the number of infrared blocks between 4 to 10, set the side length of the infrared block to vary between 6\% to 16\% of the width of the bounding box. Based on the pre-selected number of infrared blocks and the side lengths, we conduct digital attacks.

\begin{table*}
	\centering
    
    \caption{Ablation results (\%) ($k$: the number of infrared blocks; $L$: the side length of infrared blocks (\%)).}.
    \label{Table1}
	\begin{tabular}{cccccccccccccccc}
		\hline
		\multirow{2}*{$k$} & \multirow{2}*{method} & \multicolumn{2}{c}{$L=6$} & \multicolumn{2}{c}{$L=8$} & \multicolumn{2}{c}{$L=10$} & \multicolumn{2}{c}{$L=12$} & \multicolumn{2}{c}{$L=14$} & \multicolumn{2}{c}{$L=16$} & \multicolumn{2}{c}{Average}\\
		\cmidrule(r){3-4}
        \cmidrule(r){5-6}
        \cmidrule(r){7-8}
        \cmidrule(r){9-10}
        \cmidrule(r){11-12}
        \cmidrule(r){13-14}
        \cmidrule(r){15-16}
		~ & ~ & AP & ASR & AP & ASR & AP & ASR & AP & ASR & AP & ASR & AP & ASR & AP & ASR \\
		\hline

        \multirow{2}*{$4$} & R &92.4&6.1&91.9&9.2&89.4&17.4&88.2&32.5&87.4&36.1&83.5&43.9&88.8&24.2\\
        \cmidrule(r){2-16}
        ~ & AdvIB &89.3&9.9&84.8&12.2&81.1&25.4&74.2&38.7&70.2&44.8&68.3&50.8&\textbf{78.0}&\textbf{30.3}\\
        \hline

        \multirow{2}*{$5$} & R &90.5&11.6&86.8&15.8&82.7&26.8&76.9&37.9&64.7&43.0&58.3&53.1&76.7&31.4\\
        \cmidrule(r){2-16}
        ~ & AdvIB &85.6&15.5&78.7&17.1&70.4&29.3&67.5&42.0&59.7&51.9&51.5&56.9&\textbf{68.9}&\textbf{35.5}\\
        \hline

        \multirow{2}*{$6$} & R &88.1&13.9&85.3&18.2&78.7&28.5&69.9&40.7&63.4&48.3&56.2&60.1&73.6&35.0\\
        \cmidrule(r){2-16}
        ~ & AdvIB &80.3&16.6&76.8&21.0&65.3&33.1&52.6&45.9&47.7&56.9&35.4&65.2&\textbf{59.7}&\textbf{39.8}\\
        \hline

        \multirow{2}*{$7$} & R &80.9&15.5&74.7&20.3&65.8&34.0&57.4&44.6&43.7&57.4&39.3&64.8&60.3&39.4\\
        \cmidrule(r){2-16}
        ~ & AdvIB &73.7&18.2&65.4&24.9&60.9&37.6&\textbf{39.8}&\textbf{49.2}&25.2&63.5&24.1&68.5&\textbf{48.2}&\textbf{43.6}\\
        \hline

        \multirow{2}*{$8$} & R &80.1&17.7&70.8&25.1&62.2&36.9&55.8&50.3&41.5&61.8&33.7&67.4&57.4&43.2\\
        \cmidrule(r){2-16}
        ~ & AdvIB &71.7&19.9&66.3&28.2&48.2&41.4&35.3&55.2&19.7&66.9&10.6&71.3&\textbf{42.0}&\textbf{47.1}\\
        \hline

        \multirow{2}*{$9$} & R &75.7&20.3&68.4&27.8&59.7&43.5&52.3&56.0&40.9&64.1&25.4&70.6&53.7&47.1\\
        \cmidrule(r){2-16}
        ~ & AdvIB &68.3&23.2&54.2&34.8&38.9&47.0&34.6&60.8&19.2&67.4&8.3&74.0&\textbf{37.3}&\textbf{51.2}\\
        \hline

        \multirow{2}*{$10$} & R &73.2&22.3&65.4&31.4&54.6&44.5&47.3&61.2&36.6&65.6&21.8&74.2&49.1&49.9\\
        \cmidrule(r){2-16}
        ~ & AdvIB &67.3&24.9&52.7&33.1&31.7&48.6&20.2&63.5&18.4&71.3&10.9&76.8&\textbf{33.5}&\textbf{53.0}\\
        \hline
        
	\end{tabular}
\end{table*}

\begin{figure}
\centering
\includegraphics[width=0.8\columnwidth]{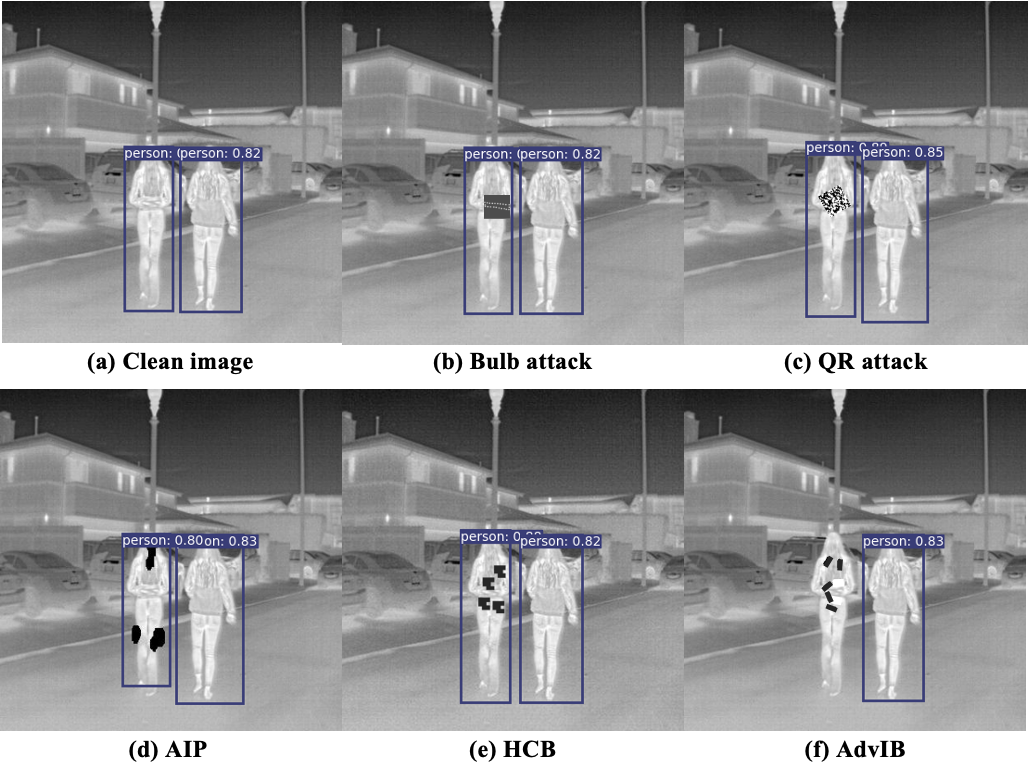} 
\caption{Example results of digital attacks. The bounding boxes indicate the infrared pedestrian detector correctly detects the person.}.
\vspace{-0.5cm}
\label{figure5}
\end{figure}

The outcomes of our experiments are presented in Table \ref{Table1}. Based on a comprehensive analysis of all experimental setups and results, we reach three main conclusions. Firstly, the efficacy of AdvIB in a digital environment outperforms that of random blocks attack (R), thereby validating the feasibility and effectiveness of our approach. Secondly, we observe a consistent increase in ASR as the number and side length of infrared blocks increase, which is a reasonable outcome in light of our expectations. Thirdly, our results indicate that when values of $k$ and $L$ are set to 7 and 12, respectively, AdvIB reduces the AP to 39.8\% while achieving an ASR of 49.2\%, demonstrating the physical feasibility of our approach in a practical configuration. We employ this specific configuration for all subsequent experiments. Furthermore, to illustrate the stealthiness of our approach, Figure \ref{figure5} (f) presents an adversarial sample generated by AdvIB. Importantly, our perturbations do not exceed those of other attacks, such as the Bulb attack \cite{ref1} (Figure \ref{figure5} (b)), the QR attack \cite{ref2} (Figure \ref{figure5} (c)), AIP \cite{CVPR2023} (Figure \ref{figure5} (d)) and HCB \cite{ref3} (Figure \ref{figure5} (e)). These findings confirm that AdvIB is capable of achieving successful digital attacks while maintaining reasonable levels of perturbations.

\textbf{Physical test.} In order to comprehensively evaluate the physical effectiveness of our approach, we conduct physical attacks at various distances and angles. The distances vary from 3.8m to 8.6m, with intervals of 1.6m, while the angles range from ${0}^{\circ}$ to ${180}^{\circ}$, with intervals of ${30}^{\circ}$.   To ensure the consistency and fairness of the physical tests, we utilize the configuration that we deemed reasonable based on the results of our previous experiments. Specifically, we use 7 infrared blocks, with a length equivalent to 12\% of the width of the bounding box. 

\begin{figure}
\centering
\includegraphics[width=0.8\columnwidth]{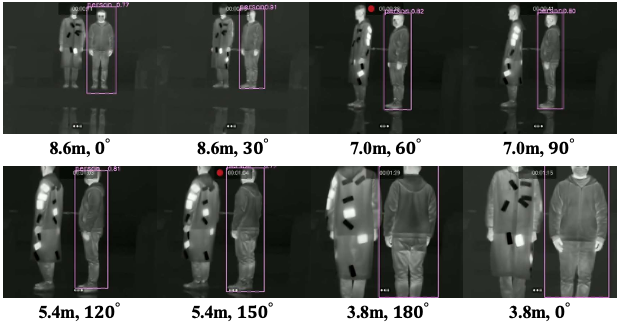} 
\caption{Example results of physical attacks.}.
\vspace{-0.5cm}
\label{figure6}
\end{figure}

\begin{table*} 
	\centering
    \caption{Physical tests at different distances and angles (\%).}.
    \label{Table2}
	\begin{tabular}{ccccccccccccccc}

        \hline
		\multirow{2}*{Distance} & \multicolumn{2}{c}{${0}^{\circ}$} & \multicolumn{2}{c}{${30}^{\circ}$} & \multicolumn{2}{c}{${60}^{\circ}$} & \multicolumn{2}{c}{${90}^{\circ}$} & \multicolumn{2}{c}{${120}^{\circ}$} & \multicolumn{2}{c}{${150}^{\circ}$} & \multicolumn{2}{c}{${180}^{\circ}$}\\
		\cline{2-15}
		~ & AP & ASR & AP & ASR & AP & ASR & AP & ASR & AP & ASR & AP & ASR & AP & ASR \\
		\hline
        
        3.8m&58.7&100&35.3&100&23.3&100&52.5&100&51.5&100&44.6&100&65.9&100\\
        \hline
        5.4m&71.2&80.6&75.0&100&52.4&45.8&34.0&81.3&54.1&40.8&64.4&97.9&41.9&100\\
        \hline
        7.0m&77.2&91.2&51.9&72.7&49.0&77.8&62.5&100&85.7&92.1&38.3&100&19.2&100\\
        \hline
        8.6m&53.4&100&32.3&100&48.8&70.7&90.5&53.4&100&86.4&76.6&48.9&73.6&91.2\\
        \hline

    \end{tabular}
\end{table*}

The results of our physical tests are presented in Table \ref{Table2}. Our comprehensive analysis of these experiments led us to draw the following conclusions: \textbf{1)} In most cases, our approach achieves a remarkable ASR of 100\%, such as 3.8m + ${0}^{\circ}$, and 7m + ${150}^{\circ}$. We also find that when the distance is 3.8m, all angles achieves 100\% ASR, with AdvIB minimizing the average AP of the model. As the distance increases, we observe a slight decrease in ASR due to the decrease in perturbations with the increase in distance. \textbf{2)} Our approach demonstrate superior ASR performance when the angles are ${0}^{\circ}$, ${30}^{\circ}$, and ${180}^{\circ}$, which almost completely paralyzes the infrared pedestrian detector. \textbf{3)} At a distance of 7m and an angle of ${180}^{\circ}$, our approach exhibites the strongest adversarial effect, reducing the AP to 19.2\% and achieving 100\% ASR, highlighting that our method has the strongest physical adversarial effect in this scene. To visually demonstrate the effectiveness of our approach, Figure \ref{figure6} displays physical samples generated by our approach that effectively attacked the infrared pedestrian detector at various distances and angles. Although the ASR remaines the same or decreases as the distance increases, our approach successfully performes physical attacks at all distances, with AdvIB achieving a physical ASR of 40.8\% in the worst-case scenario. See Supplementary Material for the video demo and more physical samples.

In summary, the experimental results presented in Table \ref{Table1} confirm the effectiveness of our proposed AdvIB method in the digital environment, and guide the selection of a reasonable attack configuration. The results shown in Table \ref{Table2} demonstrate the physical effectiveness of AdvIB at different distances and angles, and indicate that our method can achieve 100\% ASR in several cases. These findings collectively validate the effectiveness of AdvIB.  

\subsection{Evaluation of stealthiness}
\begin{figure}
\setlength{\abovecaptionskip}{-0.01cm}
\centering
\includegraphics[width=0.8\columnwidth]{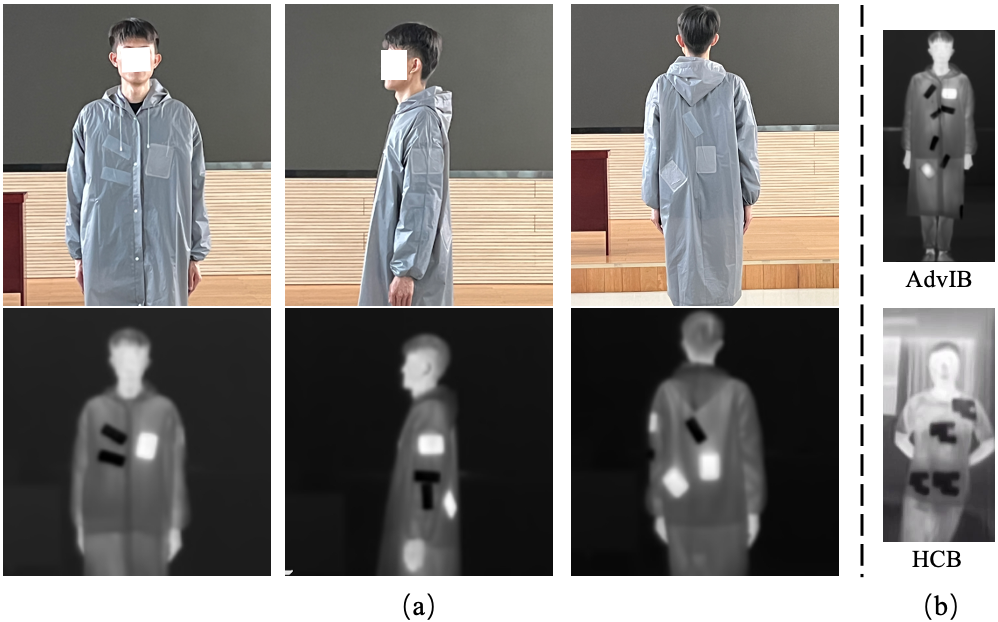} 
\caption{(a) Visual comparison between infrared image and visible light image; (b) Comparison of physical samples generated by AdvIB and HCB.}.
\vspace{-0.5cm}
\label{figure7}
\end{figure}

Figure \ref{figure5} displays digital samples generated by our approach and baselines. Our analysis indicates that the total perturbations generated by AdvIB do not exceed than those generated by the baseline methods, including Bulb attack \cite{ref1}, QR attack \cite{ref2}, AIP \cite{CVPR2023} and HCB \cite{ref3}. This finding demonstrates the stealthiness of AdvIB in the digital environment. In the physical domain, Figure \ref{figure7} (a) shows that our approach embeds the perturbations inside clothing, rendering them invisible to the naked eye without the assistance of an infrared camera. This endows AdvIB with better physical stealth than other baseline attacks, such as the Bulb attack, where a bulb plate is manually held by an individual, the QR attack, which needs an infrared suit worn on the body, and AIP, which require the infrared patch to be affixed externally to the clothing. Moreover, Figure \ref{figure7} (b) illustrates that AdvIB deploys fewer physical perturbations than HCB (7 blocks by AdvIB and 12 blocks by HCB), further demonstrating the improved stealthiness of our approach over HCB. Thus, our approach demonstrates better stealthiness than the baseline methods in both the digital and physical domains, making it a promising method for practical applications in stealthy adversarial attacks.

\begin{table}

	\centering
    \caption{Evaluation across various detectors.}.
    \label{Table4}
	\begin{tabular}{cccc}
\hline
$f$&AP(\%)&ASR(\%)&Query\\
\hline
Yolo v3&39.8&49.2&472.8\\
\hline
DETR&68.4&32.1&724.2\\
\hline
Mask Rcnn&59.7&43.4&631.6\\
\hline
Faster Rcnn&53.9&45.1&584.7\\
\hline
Libra Rcnn&42.4&50.8&518.9\\
\hline
RetinaNet&23.1&78.8&346.4\\
\hline
Average&47.9&49.9&546.4\\
\hline

\end{tabular}
\end{table}

\begin{table} 
	\centering
    \caption{Comparison of experimental results between AdvIB and HCB (In the physical tests, we use the ASR at ${0}^{\circ}$ to make a fair comparison.).}.
    \label{Table3}
	\begin{tabular}{cccc|ccc}

    \hline
		\multirow{2}*{Method} &\multirow{2}*{$k$}& \multicolumn{2}{c}{Digital(\%)} & \multicolumn{3}{c}{Physical(\%)}\\
		\cline{3-7}
		~ &  & AP & ASR & AP & ASR & Multi-angle \\
		\hline
        HCB&12&44.9&32.2&26.3&64.8&\ding{55}\\
        \hline
        AdvIB&7&\textbf{52.1}&\textbf{49.9}&\textbf{34.9}&\textbf{93.0}&\ding{51}\\
        \hline

\end{tabular}
\end{table}

\begin{table*}
	\centering
    \caption{Physical transferability of AdvIB (\%).}.
    \label{Table5}
	\begin{tabular}{ccccccccccccccc}

        \hline
		\multirow{2}*{Distance} & \multicolumn{2}{c}{${0}^{\circ}$} & \multicolumn{2}{c}{${30}^{\circ}$} & \multicolumn{2}{c}{${60}^{\circ}$} & \multicolumn{2}{c}{${90}^{\circ}$} & \multicolumn{2}{c}{${120}^{\circ}$} & \multicolumn{2}{c}{${150}^{\circ}$} & \multicolumn{2}{c}{${180}^{\circ}$}\\
		\cline{2-15}
		~ & AP & ASR & AP & ASR & AP & ASR & AP & ASR & AP & ASR & AP & ASR & AP & ASR \\
		\hline
        
        3.8m&0&100 &0 &100 &0 &100 &0 &100 &0 &100 &0 &100 &15.0 &87.5 \\
        \hline
        5.4m&21.6&80.4 &18.9 &73.0 &47.6 &57.1 &0 &100 &9.7 &93.5 &60.0 &42.0 &18.9 &83.0 \\
        \hline
        7.0m&40.0&65.0 &59.3 &43.2 &54.3 &88.6 &59.5 &42.9 &11.6 &90.7 &10.0 &92.5 &24.2 &79.0 \\
        \hline
        8.6m&23.1&84.6 &47.5 &67.2 &21.2 &81.8 &52.8 &58.3 &66.7 &71.8 &33.3 &100 &18.5 &92.6 \\
        \hline

    \end{tabular}
\end{table*}

\subsection{Evaluation of robustness}


We conduct experiments to evaluate the effectiveness of our proposed AdvIB method in attacking advanced object detectors in a black-box setting.  Specifically, we target five widely used detectors, including DETR \cite{ref43}, Mask Rcnn \cite{ref46}, Faster Rcnn \cite{ref45}, Libra Rcnn \cite{ref47} and RetinaNet \cite{ref44}. We fine-tune these pre-trained models on our proposed FLIR-IB dataset, and achieve AP scores of 91.2\%, 89.5\%, 90.8\%, 88.0\% and 93.0\% on the test set, respectively.  The evaluation metrics we used are AP, ASR, and query counts. Table \ref{Table4} presents detailed results of our experiments. Our results indicate that AdvIB significantly decreased the AP of these models by 52.1\% on average and achieved an ASR of 49.9\% on average. In HCB \cite{ref3}, AP is dropped by 44.9\%, achieving an average ASR of 32.2\%.  Therefore, our method is more efficient and robust than HCB. Interestingly, we observe that DETR is more resilient to our attack, as it showes a higher AP and lower ASR compared to the other models. We speculate that the transformer-based architecture of DETR may make it more robust against adversarial attacks. However, overall, our AdvIB method demonstrate robustness against most of the detectors we tested. To summarize, our findings suggest that AdvIB is an effective and robust method for attacking state-of-the-art object detectors. 

Table \ref{Table3} provides a summary of the experimental results obtained from our proposed method and HCB \cite{ref3}. Our findings indicate that AdvIB achieves better performance in terms of reducing the average AP of the model and achieving higher ASR in both digital and physical environments, with less perturbations compared to HCB. This indicates that AdvIB is more robust than HCB, which verifies the robustness of our method.

\subsection{Transferability of AdvIB.}
Here, we test the transfer attack of AdvIB. In the digital domain, we use the adversarial samples generated by AdvIB that could successfully attack Yolo v3 \cite{ref48} as adversarial dataset to attack advanced detectors, these detectors include Faster Rcnn \cite{ref45}, DETR \cite{ref43}, Libra Rcnn \cite{ref47}, RetinaNet \cite{ref44}, Mask Rcnn \cite{ref46}, It achieved 32.4\%, 19.9\%, 34.9\%, 58.8\%, and 29.6\% ASR, respectively. In addition, we use Yolo v3, DETR, Faster Rcnn and RetinaNet to carry out ensemble attacks on Libra Rcnn and Mask Rcnn, achieving ASR of 78.7\% and 71.3\% respectively, which shows that ensemble attacks can achieve better attack transferability. In the physical domain, Table \ref{Table5} shows the experimental results of physical transfer attacks, here, we use the physical samples generated by AdvIB that can successfully attack Yolo v3 as the adversarial dataset to attack Faster Rcnn. We can draw the following conclusions: \textbf{1)} As a whole, our method achieves effective and superior attack mobility at various distances and angles; \textbf{2)} The samples with a distance of 3.8m show the strongest adversarial transfer, achieving 87.5\% ASR at ${180}^{\circ}$ and 100\% ASR at all other angles. When the distance increases, the average ASR decreases somewhat. More results of transfer attacks are provided in the Supplementary Material.

\subsection{Ablation of pixel value}

\begin{figure}
\centering
\includegraphics[width=0.9\columnwidth]{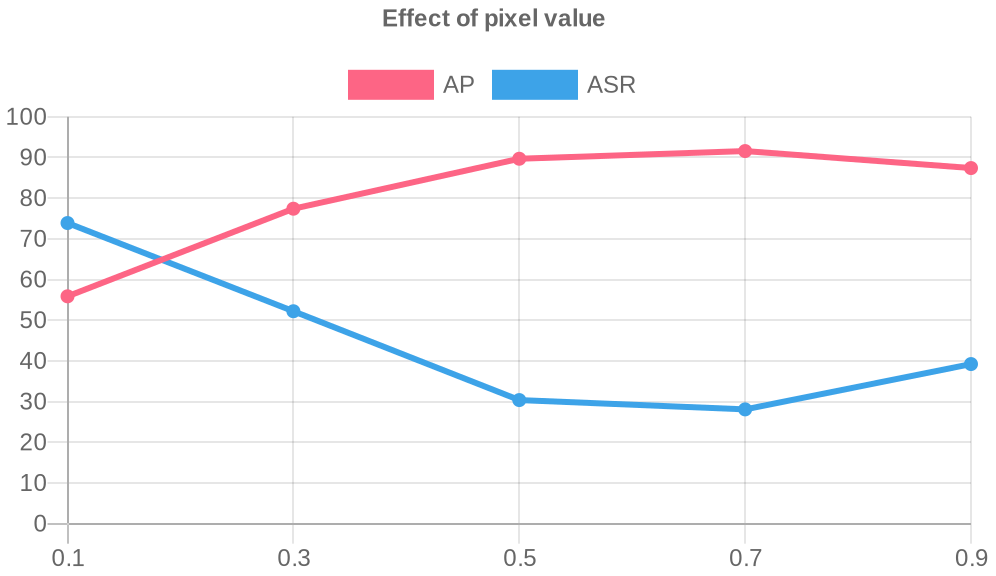} 
\caption{Ablation of pixel value.}.
\vspace{-0.6cm}
\label{figure8}
\end{figure}

In this section, we conduct experiments to analyze the effect of different pixel values ($C$) on the performance of AdvIB. Specifically, we vary $C$ between 0.1 and 0.9 in increments of 0.2 and evaluate the resulting ASR and AP. Our results, presented in Figure \ref{figure8}, demonstrate that AdvIB is most effective when $C=0.1$, achieving the highest ASR of 73.89\% and the lowest AP of 55.88\%. In contrast, when $C=0.7$, the method yields the lowest ASR of 28.09\% and the highest AP of 91.58\%. We attribute this to the fact that at $C=0.7$, the color of the infrared block closely resembles that of pedestrians in the dataset, resulting in less noticeable perturbations and weaker adversarial effects. Overall, our findings demonstrate the importance of carefully selecting the value of $C$ to optimize the effectiveness of AdvIB.

\subsection{Defense strategy discussion}
In addition to our previous in-depth analysis and discussion of AdvIB's effectiveness, stealthiness, and robustness, we now turn our attention to defense strategies against AdvIB attacks. Adversarial training \cite{ref49} is a common and effective defense technique that fundamentally enhances the model's resistance to adversarial attacks. We apply adversarial defense against AdvIB by introducing random adversarial infrared blocks into the clean sample dataset $X$ and fine-tuning the Yolo v3 model. The resulting adversarial training model achieves 91.3\% and 95.4\% AP on the training and test sets, respectively. Using AdvIB, we launch digital attacks on the adversarial training model and measure its ASR, Query, and AP, which are 14.9\%, 901.6, and 80.7\%, respectively (compared to 49.2\%, 472.8, and 39.8\%, respectively, before adversarial training).  Our findings indicate that while the adversarial training model does enhance robustness against AdvIB attacks, it cannot guarantee complete defense.

\section{Conclusion}
In this paper, we present AdvIB, a novel black-box physical attack that introduces hot and cold blocks as physical perturbations to generate the most adversarial physical parameters via DE optimization, resulting in improved effectiveness, stealthiness, and robustness of infrared attack. Our approach is physically feasible and stealthy, as the perturbations can be placed inside clothing, making them difficult for human observers to detect without the aid of an infrared camera. We conduct comprehensive experimental evaluations to demonstrate the efficacy of AdvIB in evading infrared pedestrian detectors in both digital and physical environments. Our method is more efficient than existing baseline methods and can be easily replicated in practical scenarios. Overall, AdvIB represents a significant advancement in infrared attacks, with its unique combination of physical perturbations, optimization techniques, and stealthy nature.

Our approach presents novel ideas for future attacks on thermal infrared detectors by embedding perturbations inside the target object, rather than exposing them externally. This makes the physical attack stealthy and capable of deceiving the detectors. Moreover, comprehensive defense strategies against physical attacks on infrared imaging systems will be a critical research focus in the future.

\bibliographystyle{ieeetr}

\bibliography{IEEEfull}

\begin{thebibliography}{10}

\bibitem{ref1}
X.~Zhu, X.~Li, J.~Li, Z.~Wang, and X.~Hu, ``Fooling thermal infrared pedestrian
  detectors in real world using small bulbs,'' in {\em Proceedings of the AAAI
  Conference on Artificial Intelligence}, vol.~35, pp.~3616--3624, 2021.

\bibitem{ref2}
X.~Zhu, Z.~Hu, S.~Huang, J.~Li, and X.~Hu, ``Infrared invisible clothing:
  Hiding from infrared detectors at multiple angles in real world,'' in {\em
  Proceedings of the IEEE/CVF Conference on Computer Vision and Pattern
  Recognition}, pp.~13317--13326, 2022.

\bibitem{ref3}
H.~Wei, Z.~Wang, X.~Jia, Y.~Zheng, H.~Tang, S.~Satoh, and Z.~Wang, ``Hotcold
  block: Fooling thermal infrared detectors with a novel wearable design,''
  {\em arXiv preprint arXiv:2212.05709}, 2022.

\bibitem{CVPR2023}
X.~Wei, J.~Yu, and Y.~Huang, ``Physically adversarial infrared patches with
  learnable shapes and locations,'' {\em CoRR}, vol.~abs/2303.13868, 2023.

\bibitem{ref4}
C.~Szegedy, W.~Zaremba, I.~Sutskever, J.~Bruna, D.~Erhan, I.~J. Goodfellow, and
  R.~Fergus, ``Intriguing properties of neural networks,'' in {\em 2nd
  International Conference on Learning Representations, {ICLR} 2014, Banff, AB,
  Canada, April 14-16, 2014, Conference Track Proceedings} (Y.~Bengio and
  Y.~LeCun, eds.), 2014.

\bibitem{ref5}
R.~Duan, Y.~Chen, D.~Niu, Y.~Yang, A.~K. Qin, and Y.~He, ``Advdrop: Adversarial
  attack to dnns by dropping information,'' in {\em Proceedings of the IEEE/CVF
  International Conference on Computer Vision}, pp.~7506--7515, 2021.

\bibitem{ref6}
Y.~Dong, S.~Cheng, T.~Pang, H.~Su, and J.~Zhu, ``Query-efficient black-box
  adversarial attacks guided by a transfer-based prior,'' {\em IEEE
  Transactions on Pattern Analysis and Machine Intelligence}, vol.~44, no.~12,
  pp.~9536--9548, 2021.

\bibitem{ref7}
S.~Feng, F.~Feng, X.~Xu, Z.~Wang, Y.~Hu, and L.~Xie, ``Digital watermark
  perturbation for adversarial examples to fool deep neural networks,'' in {\em
  2021 International Joint Conference on Neural Networks (IJCNN)}, pp.~1--8,
  IEEE, 2021.

\bibitem{ref8}
Y.~Zhang, X.~Tian, Y.~Li, X.~Wang, and D.~Tao, ``Principal component
  adversarial example,'' {\em IEEE Transactions on Image Processing}, vol.~29,
  pp.~4804--4815, 2020.

\bibitem{ref52}
R.~Wang, F.~Juefei-Xu, Q.~Guo, Y.~Huang, X.~Xie, L.~Ma, and Y.~Liu, ``Amora:
  Black-box adversarial morphing attack,'' in {\em Proceedings of the 28th ACM
  International Conference on Multimedia}, pp.~1376--1385, 2020.

\bibitem{ref51}
X.~Han, G.~Xu, Y.~Zhou, X.~Yang, J.~Li, and T.~Zhang, ``Physical backdoor
  attacks to lane detection systems in autonomous driving,'' in {\em
  Proceedings of the 30th ACM International Conference on Multimedia},
  pp.~2957--2968, 2022.

\bibitem{ref9}
J.~Wang, A.~Liu, X.~Bai, and X.~Liu, ``Universal adversarial patch attack for
  automatic checkout using perceptual and attentional bias,'' {\em IEEE
  Transactions on Image Processing}, vol.~31, pp.~598--611, 2021.

\bibitem{ref10}
B.~G. Doan, M.~Xue, S.~Ma, E.~Abbasnejad, and D.~C. Ranasinghe, ``Tnt attacks!
  universal naturalistic adversarial patches against deep neural network
  systems,'' {\em IEEE Transactions on Information Forensics and Security},
  vol.~17, pp.~3816--3830, 2022.

\bibitem{ref53}
X.~Du and C.-M. Pun, ``Adversarial image attacks using multi-sample and
  most-likely ensemble methods,'' in {\em Proceedings of the 28th ACM
  International Conference on Multimedia}, pp.~1634--1642, 2020.

\bibitem{ref16}
W.~Wu, Y.~Su, M.~R. Lyu, and I.~King, ``Improving the transferability of
  adversarial samples with adversarial transformations,'' in {\em Proceedings
  of the IEEE/CVF conference on computer vision and pattern recognition},
  pp.~9024--9033, 2021.

\bibitem{ref17}
Y.~Dong, F.~Liao, T.~Pang, H.~Su, J.~Zhu, X.~Hu, and J.~Li, ``Boosting
  adversarial attacks with momentum,'' in {\em Proceedings of the IEEE
  conference on computer vision and pattern recognition}, pp.~9185--9193, 2018.

\bibitem{ref18}
C.~Luo, Q.~Lin, W.~Xie, B.~Wu, J.~Xie, and L.~Shen, ``Frequency-driven
  imperceptible adversarial attack on semantic similarity,'' in {\em
  Proceedings of the IEEE/CVF Conference on Computer Vision and Pattern
  Recognition}, pp.~15315--15324, 2022.

\bibitem{ref19}
R.~Duan, X.~Ma, Y.~Wang, J.~Bailey, A.~K. Qin, and Y.~Yang, ``Adversarial
  camouflage: Hiding physical-world attacks with natural styles,'' in {\em
  Proceedings of the IEEE/CVF conference on computer vision and pattern
  recognition}, pp.~1000--1008, 2020.

\bibitem{ref20}
J.~Su, D.~V. Vargas, and K.~Sakurai, ``One pixel attack for fooling deep neural
  networks,'' {\em IEEE Transactions on Evolutionary Computation}, vol.~23,
  no.~5, pp.~828--841, 2019.

\bibitem{ref21}
C.~Xie, Z.~Zhang, Y.~Zhou, S.~Bai, J.~Wang, Z.~Ren, and A.~L. Yuille,
  ``Improving transferability of adversarial examples with input diversity,''
  in {\em Proceedings of the IEEE/CVF Conference on Computer Vision and Pattern
  Recognition}, pp.~2730--2739, 2019.

\bibitem{ref22}
J.~Byun, S.~Cho, M.-J. Kwon, H.-S. Kim, and C.~Kim, ``Improving the
  transferability of targeted adversarial examples through object-based diverse
  input,'' in {\em Proceedings of the IEEE/CVF Conference on Computer Vision
  and Pattern Recognition}, pp.~15244--15253, 2022.

\bibitem{ref23}
B.~Bonnet, T.~Furon, and P.~Bas, ``Generating adversarial images in quantized
  domains,'' {\em IEEE Transactions on Information Forensics and Security},
  vol.~17, pp.~373--385, 2021.

\bibitem{ref24}
Q.~Li, Y.~Qi, Q.~Hu, S.~Qi, Y.~Lin, and J.~S. Dong, ``Adversarial adaptive
  neighborhood with feature importance-aware convex interpolation,'' {\em IEEE
  Transactions on Information Forensics and Security}, vol.~16, pp.~2447--2460,
  2020.

\bibitem{ref25}
A.~M{\k{a}}dry, A.~Makelov, L.~Schmidt, D.~Tsipras, and A.~Vladu, ``Towards
  deep learning models resistant to adversarial attacks,'' {\em stat},
  vol.~1050, p.~9, 2017.

\bibitem{ref26}
A.~S. Shamsabadi, R.~Sanchez-Matilla, and A.~Cavallaro, ``Colorfool: Semantic
  adversarial colorization,'' in {\em Proceedings of the IEEE/CVF Conference on
  Computer Vision and Pattern Recognition}, pp.~1151--1160, 2020.

\bibitem{ref27}
Z.~Zhao, Z.~Liu, and M.~Larson, ``Towards large yet imperceptible adversarial
  image perturbations with perceptual color distance,'' in {\em Proceedings of
  the IEEE/CVF Conference on Computer Vision and Pattern Recognition},
  pp.~1039--1048, 2020.

\bibitem{ref28}
Z.~Hu, S.~Huang, X.~Zhu, F.~Sun, B.~Zhang, and X.~Hu, ``Adversarial texture for
  fooling person detectors in the physical world,'' in {\em Proceedings of the
  IEEE/CVF Conference on Computer Vision and Pattern Recognition},
  pp.~13307--13316, 2022.

\bibitem{ref29}
D.~Wang, T.~Jiang, J.~Sun, W.~Zhou, Z.~Gong, X.~Zhang, W.~Yao, and X.~Chen,
  ``Fca: Learning a 3d full-coverage vehicle camouflage for multi-view physical
  adversarial attack,'' in {\em Proceedings of the AAAI Conference on
  Artificial Intelligence}, vol.~36, pp.~2414--2422, 2022.

\bibitem{ref30}
N.~Suryanto, Y.~Kim, H.~Kang, H.~T. Larasati, Y.~Yun, T.-T.-H. Le, H.~Yang,
  S.-Y. Oh, and H.~Kim, ``Dta: Physical camouflage attacks using differentiable
  transformation network,'' in {\em Proceedings of the IEEE/CVF Conference on
  Computer Vision and Pattern Recognition}, pp.~15305--15314, 2022.

\bibitem{ref31}
J.~Liu, B.~Lu, M.~Xiong, T.~Zhang, and H.~Xiong, ``Adversarial attack with
  raindrops,'' {\em arXiv preprint arXiv:2302.14267}, 2023.

\bibitem{ref32}
L.~Zhai, F.~Juefei-Xu, Q.~Guo, X.~Xie, L.~Ma, W.~Feng, S.~Qin, and Y.~Liu,
  ``Adversarial rain attack and defensive deraining for dnn perception,'' {\em
  arXiv preprint arXiv:2009.09205}, 2022.

\bibitem{ref33}
X.~Wei, Y.~Guo, and J.~Yu, ``Adversarial sticker: A stealthy attack method in
  the physical world,'' {\em IEEE Transactions on Pattern Analysis and Machine
  Intelligence}, 2022.

\bibitem{ref34}
K.~Eykholt, I.~Evtimov, E.~Fernandes, B.~Li, A.~Rahmati, C.~Xiao, A.~Prakash,
  T.~Kohno, and D.~Song, ``Robust physical-world attacks on deep learning
  visual classification,'' in {\em Proceedings of the IEEE conference on
  computer vision and pattern recognition}, pp.~1625--1634, 2018.

\bibitem{ref35}
S.~Thys, W.~Van~Ranst, and T.~Goedem{\'e}, ``Fooling automated surveillance
  cameras: adversarial patches to attack person detection,'' in {\em
  Proceedings of the IEEE/CVF conference on computer vision and pattern
  recognition workshops}, pp.~0--0, 2019.

\bibitem{ref36}
S.~Huang, X.~Liu, X.~Yang, and Z.~Zhang, ``An improved shapeshifter method of
  generating adversarial examples for physical attacks on stop signs against
  faster r-cnns,'' {\em Computers \& Security}, vol.~104, p.~102120, 2021.

\bibitem{ref37}
R.~Duan, X.~Mao, A.~K. Qin, Y.~Chen, S.~Ye, Y.~He, and Y.~Yang, ``Adversarial
  laser beam: Effective physical-world attack to dnns in a blink,'' in {\em
  Proceedings of the IEEE/CVF Conference on Computer Vision and Pattern
  Recognition}, pp.~16062--16071, 2021.

\bibitem{ref38}
Y.~Zhong, X.~Liu, D.~Zhai, J.~Jiang, and X.~Ji, ``Shadows can be dangerous:
  Stealthy and effective physical-world adversarial attack by natural
  phenomenon,'' in {\em Proceedings of the IEEE/CVF Conference on Computer
  Vision and Pattern Recognition}, pp.~15345--15354, 2022.

\bibitem{ref39}
A.~Zolfi, M.~Kravchik, Y.~Elovici, and A.~Shabtai, ``The translucent patch: A
  physical and universal attack on object detectors,'' in {\em Proceedings of
  the IEEE/CVF Conference on Computer Vision and Pattern Recognition},
  pp.~15232--15241, 2021.

\bibitem{ref40}
A.~Guesmi, M.~A. Hanif, and M.~Shafique, ``Advrain: Adversarial raindrops to
  attack camera-based smart vision systems,'' {\em arXiv preprint
  arXiv:2303.01338}, 2023.

\bibitem{PSO}
J.~Kennedy and R.~Eberhart, ``Particle swarm optimization,'' in {\em
  Proceedings of ICNN'95-international conference on neural networks}, vol.~4,
  pp.~1942--1948, IEEE, 1995.

\bibitem{ref50}
A.~Athalye, L.~Engstrom, A.~Ilyas, and K.~Kwok, ``Synthesizing robust
  adversarial examples,'' in {\em International conference on machine
  learning}, pp.~284--293, PMLR, 2018.

\bibitem{ref15}
R.~Storn and K.~Price, ``Differential evolution-a simple and efficient
  heuristic for global optimization over continuous spaces,'' {\em Journal of
  global optimization}, vol.~11, no.~4, p.~341, 1997.

\bibitem{ref48}
A.~Farhadi and J.~Redmon, ``Yolov3: An incremental improvement,'' in {\em
  Computer vision and pattern recognition}, vol.~1804, pp.~1--6, Springer
  Berlin/Heidelberg, Germany, 2018.

\bibitem{ref43}
N.~Carion, F.~Massa, G.~Synnaeve, N.~Usunier, A.~Kirillov, and S.~Zagoruyko,
  ``End-to-end object detection with transformers,'' in {\em Computer
  Vision--ECCV 2020: 16th European Conference, Glasgow, UK, August 23--28,
  2020, Proceedings, Part I 16}, pp.~213--229, Springer, 2020.

\bibitem{ref46}
K.~He, G.~Gkioxari, P.~Doll{\'a}r, and R.~Girshick, ``Mask r-cnn,'' in {\em
  Proceedings of the IEEE international conference on computer vision},
  pp.~2961--2969, 2017.

\bibitem{ref45}
S.~Ren, K.~He, R.~Girshick, and J.~Sun, ``Faster r-cnn: Towards real-time
  object detection with region proposal networks,'' {\em Advances in neural
  information processing systems}, vol.~28, 2015.

\bibitem{ref47}
J.~Pang, K.~Chen, J.~Shi, H.~Feng, W.~Ouyang, and D.~Lin, ``Libra r-cnn:
  Towards balanced learning for object detection,'' in {\em Proceedings of the
  IEEE/CVF conference on computer vision and pattern recognition},
  pp.~821--830, 2019.

\bibitem{ref44}
T.-Y. Lin, P.~Goyal, R.~Girshick, K.~He, and P.~Doll{\'a}r, ``Focal loss for
  dense object detection,'' in {\em Proceedings of the IEEE international
  conference on computer vision}, pp.~2980--2988, 2017.

\bibitem{ref49}
T.~Bai, J.~Luo, J.~Zhao, B.~Wen, and Q.~Wang, ``Recent advances in adversarial
  training for adversarial robustness,'' in {\em Proceedings of the Thirtieth
  International Joint Conference on Artificial Intelligence, {IJCAI} 2021,
  Virtual Event / Montreal, Canada, 19-27 August 2021} (Z.~Zhou, ed.),
  pp.~4312--4321, ijcai.org, 2021.

\end{thebibliography}


\end{document}